\documentclass{article}
\usepackage{dcase2021,amsmath,graphicx,url,times,booktabs, tabularx}
\usepackage[nolist,nohyperlinks]{acronym}
\usepackage{amssymb}
\usepackage{caption}
\usepackage{cite}

\makeatletter
\def\bstctlcite{\@ifnextchar[{\@bstctlcite}{\@bstctlcite[@auxout]}}
\def\@bstctlcite[#1]#2{\@bsphack
  \@for\@citeb:=#2\do{%
    \edef\@citeb{\expandafter\@firstofone\@citeb}%
    \if@filesw\immediate\write\csname #1\endcsname{\string\citation{\@citeb}}\fi}%
  \@esphack}
\makeatother 

\title{Evaluating Off-the-Shelf Machine Listening and Natural Language Models for Automated Audio Captioning}

\name{Benno Weck$^{1,2}$,
      Xavier Favory$^{2}$,
      Konstantinos Drossos$^{3}$,
      Xavier Serra$^{2}$
      }
\address{$^1$ Huawei Technologies, Munich Research Center, Germany\\
        \{firstname.lastname\}@huawei.com\\          
        $^2$ Music Technology Group, Universitat Pompeu Fabra, Spain\\
        benno.weck01@estudiant.upf.edu,\{firstname.lastname\}@upf.edu\\ 
        $^3$ Audio Research Group, Tampere University, Finland\\
        \{firstname.lastname\}@tuni.fi\\
 }

\begin{document}

\ninept
\maketitle
\bstctlcite{IEEEexample:BSTcontrol}

\begin{abstract}
\Ac{AAC} is the task of automatically generating textual descriptions for general audio signals.
A captioning system has to identify various information from the input signal and express it with natural language.
Existing works mainly focus on investigating new methods and try to improve their performance measured on existing datasets.
Having attracted attention only recently, very few works on \ac{AAC} study the performance of existing pre-trained audio and natural language processing resources.
In this paper, we evaluate the performance of off-the-shelf models with a Transformer-based captioning approach.
We utilize the freely available Clotho dataset to compare four different pre-trained machine listening models, four word embedding models, and their combinations in many different settings.
Our evaluation suggests that YAMNet combined with BERT embeddings produces the best captions.
Moreover, in general, fine-tuning pre-trained word embeddings can lead to better performance.
Finally, we show that sequences of audio embeddings can be processed using a Transformer encoder to produce higher-quality captions.
\end{abstract}

\begin{keywords}
audio captioning, transfer learning, word embeddings, machine listening, transformer
\end{keywords}

\acresetall

\section{Introduction}
\label{sec:intro}
\Ac{AAC} is an inter-modal translation task, where existing methods take an audio signal as input and generate a textual description, i.e. a caption, of its contents~\cite{drossos:2017:waspaa}.
The generated captions contain information about various aspects of the content of the audio signal, ranging from identification of sound events to knowledge about spatiotemporal interactions, foreground and background disambiguation, surroundings, textures, and other high-level information~\cite{xu:2021:ICASSP:02,eren:2020:ism,drossos:2019:dcase_b}.

To our knowledge, all published works focusing on \ac{AAC} solely employ deep learning methods~\cite{xu:2021:ICASSP:02, koizumi2020audio, Koizumi2020TRACKE, eren:2020:ism, eren:2020:arxiv, chen:2020:dcase, Nguyen2020TemporalSO, Tran2020WaveTransformerAN, akir2020MultitaskRB, xu:2020:dcase, takeuchi:2020:dcase, mengyue:2019:icasssp, ikawa:2019:dcase, xu:2021:ISCSLP, drossos:2017:waspaa, kim:2019:audiocaps}.
Most of them follow an encoder-decoder scheme and address the task as a \ac{seq2seq} learning problem~\cite{Sutskever2014SequenceTS}. 
\Ac{CNN}-based encoders are often utilized, for example, in ~\cite{xu:2021:ICASSP:02, eren:2020:arxiv, eren:2020:ism, xu:2020:dcase, chen:2020:dcase, kim:2019:audiocaps}, and 
\ac{RNN}-based decoders can be used in order to generate the captions~\cite{xu:2021:ICASSP:02, drossos:2017:waspaa, Nguyen2020TemporalSO, akir2020MultitaskRB, eren:2020:arxiv, eren:2020:ism, xu:2020:dcase, takeuchi:2020:dcase, xu:2021:ISCSLP, ikawa:2019:dcase, kim:2019:audiocaps, mengyue:2019:icasssp}.
Recently, more and more methods have involved an attention mechanism.
For example, as a technique to enable the decoder to focus only on certain parts of the latent representation extracted by the encoder~\cite{xu:2021:ICASSP:02, kim:2019:audiocaps, drossos:2017:waspaa}.
Or more generally, other approaches~\cite{Koizumi2020TRACKE, Tran2020WaveTransformerAN, wuyusong2020_t6, chen:2020:dcase, yuan2021_t6} employ a Transformer model~\cite{Vaswani2017AttentionIA}.
This type of model seems particularly adequate for \ac{AAC} since it led to groundbreaking results in multiple fields, such as \ac{NLP}, computer vision, and audio processing~\cite{Lin2021TransformerSurvey}.

Transfer learning is a popular technique often employed in \ac{NLP} and \ac{MaL} tasks.
However, existing approaches in \ac{AAC} often do not take advantage of any pre-trained resources and instead train their models from scratch.
Only recently, a few published papers \cite{xu:2021:ICASSP:02, Koizumi2020TRACKE, koizumi2020audio, eren:2020:ism} propose to incorporate pre-trained audio models such as VGGish~\cite{audiosetmodels} or to rely on word embedding models such as word2vec~\cite{Mikolov2013DistributedRO}.
Given the large number of available pre-trained models in \ac{MaL} and \ac{NLP}, it is still unclear which models are most suited for \ac{AAC}. 
Moreover, incorporating these models into a Transformer-based \ac{AAC} system can involve some specific design choices that are also yet overlooked.

In this paper, we focus on investigating the use of pre-trained models taken from \ac{MaL} and \ac{NLP} in the context of \ac{AAC}. 
In particular, we are interested in identifying which available resources are the most valuable and how to combine them efficiently in a Transformer-based \ac{AAC} system.
We use various off-the-shelf pre-trained audio and word encoding methods.
Our contributions are:

\begin{itemize}
    \item We adapt a Transformer-based \ac{AAC} method that can use different pre-trained \ac{MaL} and \ac{NLP} models,
    \item We conduct a thorough investigation of the performance of our method by combining pre-trained models in various settings,
    \item We identify what combinations of techniques make pre-trained resources specifically beneficial for an \ac{AAC} system. 
    We consider fine-tuning word embeddings, using an adapter to process audio embeddings, and the usage of overlap when extracting audio embeddings.
\end{itemize}

The rest of the paper is structured as follows: In Section~\ref{sec:method} we present our method and in Section~\ref{sec:evaluation} we outline the evaluation process. The results are presented and discussed in Section~\ref{sec:results}.
Finally, Section~\ref{sec:conclusions} concludes the paper. 
\section{Method}
\label{sec:method}
\begin{figure}[t]
  \centering
  \includegraphics[trim=160 5 190 5,clip,width=.7\columnwidth]{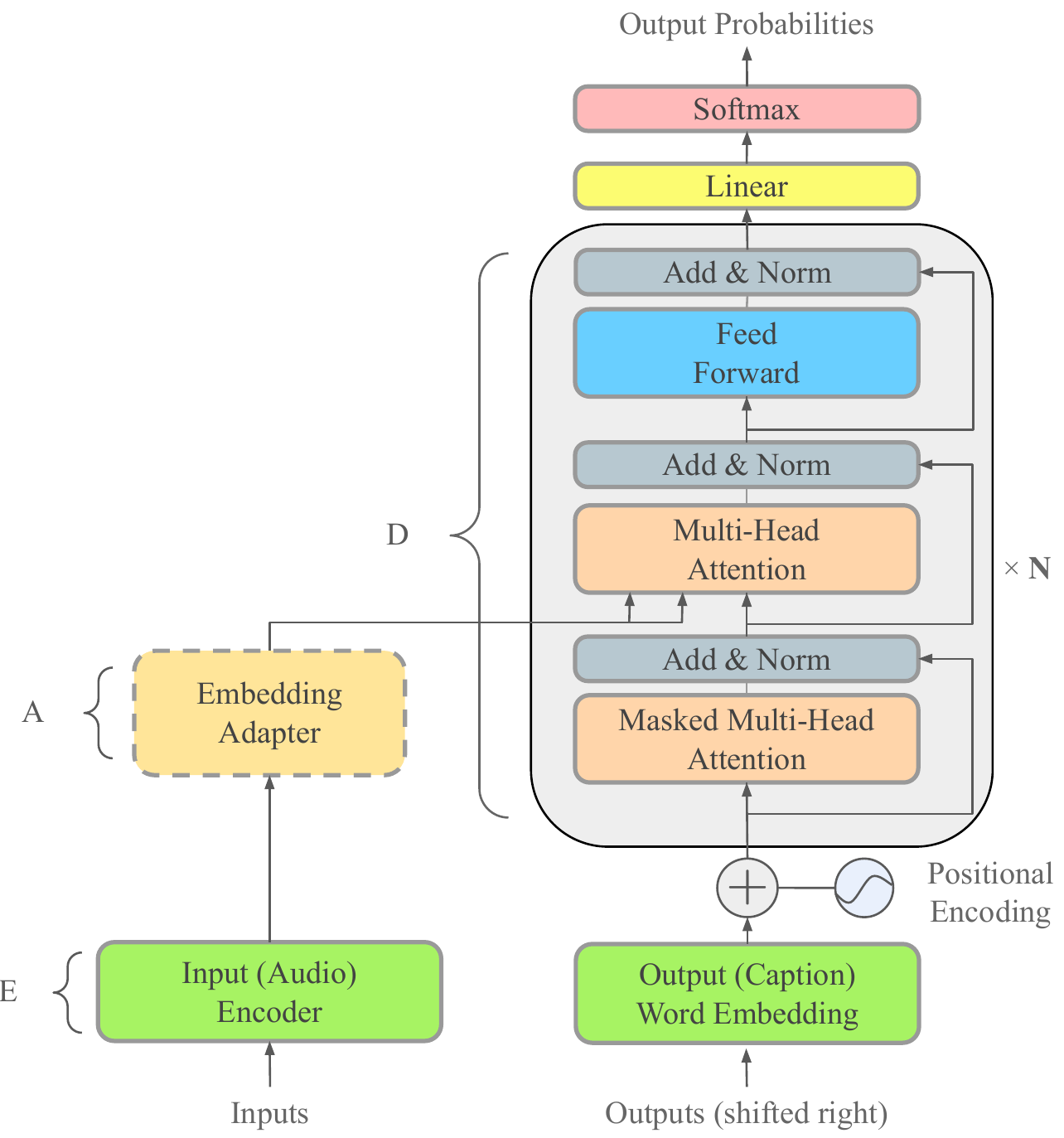}
  \caption{Model architecture.}
  \label{fig:model}
\end{figure}

In our study, we adopt a Transformer-based model architecture which has been shown to produce state-of-the-art results for \ac{AAC}~\cite{Tran2020WaveTransformerAN, wuyusong2020_t6, yuan2021_t6}.
An overview of our method is presented in Figure~\ref{fig:model}. 
It consists of an audio encoder, $\text{E}(\cdot)$, an embeddings' adapter, $\text{A}(\cdot)$, and a decoder $\text{D}(\cdot)$.
As $\text{E}$, we employ different pre-trained models for general audio processing.
For $\text{A}$, we compare the use of no adapter, a \ac{MLP}, and a \ac{MHA} component.
The output of $\text{A}$ is used together with word embeddings of the previously predicted words as an input to $\text{D}$, a Transformer-based decoder\footnote{For a complete description of the Transformer decoder, refer to~\cite{Vaswani2017AttentionIA}.}. 

A sequence of audio features $\mathbf{X}\in\mathbb{R}^{T\times F}$ with $T$ vectors of $F$ features is used as an input to the method, which outputs a sequence of one-hot encoded tokens $\mathbf{S}\in[0, 1]^{K\times W}$, where
$K$ corresponds to the number of tokens in the generated caption and $W$ to the size of the considered vocabulary.
More specifically, $\mathbf{X}$ is used as an input to $\text{E}$ as
\begin{equation}
    \mathbf{Z} = \text{E}(\mathbf{X})\text{,}
\end{equation}
\noindent
where $\mathbf{Z}\in\mathbb{R}^{T'\times F'}$ is a sequence of $T'$ intermediate representations with $F'$ features provided by the pre-trained model (i.e. an audio embedding sequence).
Then, the adapter $\text{A}$ will process $\mathbf{Z}$ as
\begin{equation}\label{eqn:adapter}
    \mathbf{Z'} = \text{A}(\mathbf{Z})\text{,}
\end{equation}
\noindent
where $\mathbf{Z'}\in\mathbb{R}^{T'\times F''}$ and $F''$ is the dimensionality of the features that $\text{A}$ outputs.
Finally, the decoder $\text{D}$ will predict the probability distribution of appearance over the $W$ words at the $k$-th step, $\mathbf{S}_{k}$, as
\begin{equation}
    \mathbf{S}_{k} = \text{D}(\mathbf{Z'}, \mathbf{S}'_{0},\ldots,\mathbf{S}'_{k-1})\text{,}
\end{equation}
\noindent
where $\mathbf{S}'_{i}$ is a learned word embedding for step $i$, and $\mathbf{S}'_{0} = \{0\}^{W'}$.
As $\mathbf{S}'$, we make use of different pre-trained \ac{NLP} models. 

We employ different audio embedding models that are optimized for a task different from \ac{AAC}.
The extracted audio embeddings might contain information that is specific to the corresponding source task and not necessarily optimal for \ac{AAC}.
We do not fine-tune the models in our experiment, but instead, we study the usage of different adapters $\text{A}$ that process the audio embeddings $\mathbf{Z}$.

The Transformer decoder D consists of $N$ blocks, each of them having two serially cascaded \ac{MHA} layers that perform self and cross-modal (i.e. between audio and words) attention, respectively.
The output of the second, cross-modal attention, is given as an input to a linear layer and a layer normalization process.
The word embeddings $\mathbf{S}'$ are used as an input to $\text{D}$.
Following the original proposal of the Transformer model, we apply a positional encoding to the input word embeddings.
To generate the captions, the decoder can be sampled until the desired caption length is met or a special token indicating the end of a sentence is produced.

\section{Evaluation}
\label{sec:evaluation}
In our study, we compare the performance of four pre-trained audio processing models, two audio embedding adapters, and four pre-trained \ac{NLP} models for \ac{AAC}, using the \ac{AAC} dataset Clotho~\cite{Drossos2020ClothoAA}. 

\subsection{Dataset, metrics, and experiments}
\label{ssec:dataset}
The Clotho dataset contains a training, a validation, and an evaluation split, comprising 3839, 1045, and 1045 audio examples, respectively.
Each audio example is annotated with five captions, and we consider one audio-caption pair a single training example.
The performance is assessed using the SPIDEr score~\cite{Liu2017SPIDEr}.
This metric is widely established in the community (e.g. in the \ac{AAC} task from the 2021 DCASE Challenge\footnote{\url{http://dcase.community/challenge2021/task-automatic-audio-captioning}}) as it highly correlates with the human judgment of caption quality~\cite{Liu2017SPIDEr}.

In all our experiments, the decoder part of our model consists of $N=3$ stacked Transformer decoder blocks with four 128-dimensional attention heads each, similar to what is used in \cite{Tran2020WaveTransformerAN,wuyusong2020_t6}.
During training, all models are optimized by minimizing the cross-entropy loss between the predicted sentence and the target caption using the Adam algorithm ($\alpha=0.001, \beta_{1}=0.9, \beta_{2}=0.999,$ and $\epsilon=10^{-8}$) with a minibatch size of 256.
Early stopping is applied after ten epochs with no improvement of the loss calculated on the held-out validation set.
The best model according to this loss is then evaluated on the evaluation set.

To avoid bias in the results, we repeat all experiments ten times with different random initialization and report the mean statistic for all scores.
We test all combinations of encoder models with different overlaps, audio embedding adapter layers, fixed or fine-tuned word embeddings.
In total, it constitutes 264 different settings.

\subsection{Audio embedding models}
\label{ssec:sourcemodelcomp}
For each employed audio embedding model, we follow the authors' methodology to extract audio embeddings using their pre-trained models.
We use the pre-trained models as encoders with frozen weights, i.e. without fine-tuning. 

Existing \ac{AAC} approaches use audio encoders with different hop-sizes.
In this work, we study the impact of using overlap when extracting the audio embeddings. 
More specifically, we use two different settings for the embedding extraction hop-size, corresponding to 50\% overlap and no overlap.

\begin{table}[h!]
\centering
\caption{Audio encoder models compared in this study.
}
\label{table:audiosourcemodels}
\begin{tabular}{c c c c c} 
  Encoder & Dimensionality & Window & Learning\\ 
 \midrule
 VGGish & 128 & 0.96 s & supervised\\ 
 YAMNet & 1024 & 0.96 s & supervised \\
 OpenL3 & 512 & 1.00 s & self-supervised\\
 COALA & 1152 & 2.20 s & contrastive \\
\end{tabular}
\end{table}

Table~\ref{table:audiosourcemodels} gives an overview of the audio source models we compared.
All four are \ac{CNN}-based models.
The first, \textbf{VGGish}~\cite{audiosetmodels}, is inspired by the VGG architecture mainly used in computer vision~\cite{Simonyan2015VGG}.
It was trained on a preliminary version of the YouTube-8M dataset in a supervised fashion~\cite{YouTube8M}.
It extracts 128-dimensional embeddings from roughly 1 second of audio. 
The second audio embedding model is \textbf{YAMNet}, which also draws inspiration from computer vision models~\cite{audiosetmodels}.
It employs a MobileNet~\cite{howard2017mobilenets} architecture to extract embeddings with dimensionality 1024 from almost 1 second of audio.
The model was trained to predict 521 audio event classes on the AudioSet dataset~\cite{gemmeke:2017:icassp}.
The third model, \textbf{OpenL3}~\cite{Cramer2019openl3}, is a modified and freely available version of the $L^3$-Net~\cite{arandjelovic2017look}.
OpenL3 is trained in a self-supervised way in an audio-visual correspondence task, relying on videos from AudioSet.
From the multiple variants that the authors provide, we chose the model configuration that produces an embedding of size 512, and that was trained with 128 Mel bands as input representation in the environmental sound setting. 
The fourth and final model is \textbf{COALA}~\cite{Favory2020COALA}, a model trained by taking advantage of user-provided tags in Freesound\footnote{The training data from COALA and Clotho are disjoint sets.}~\cite{freesoundACMMM13}.
During training, it employed a contrastive learning approach to align audio and associated tag embeddings, producing an audio embedding model that can extract semantically enriched audio representations.
The model produces embeddings from 2.2-second patches.

\subsection{Adapter Layers}
\label{ssec:adapterlayers}
\begin{figure}[t]
  \centering
  \includegraphics[trim=100 40 50 140,clip,width=.85\columnwidth]{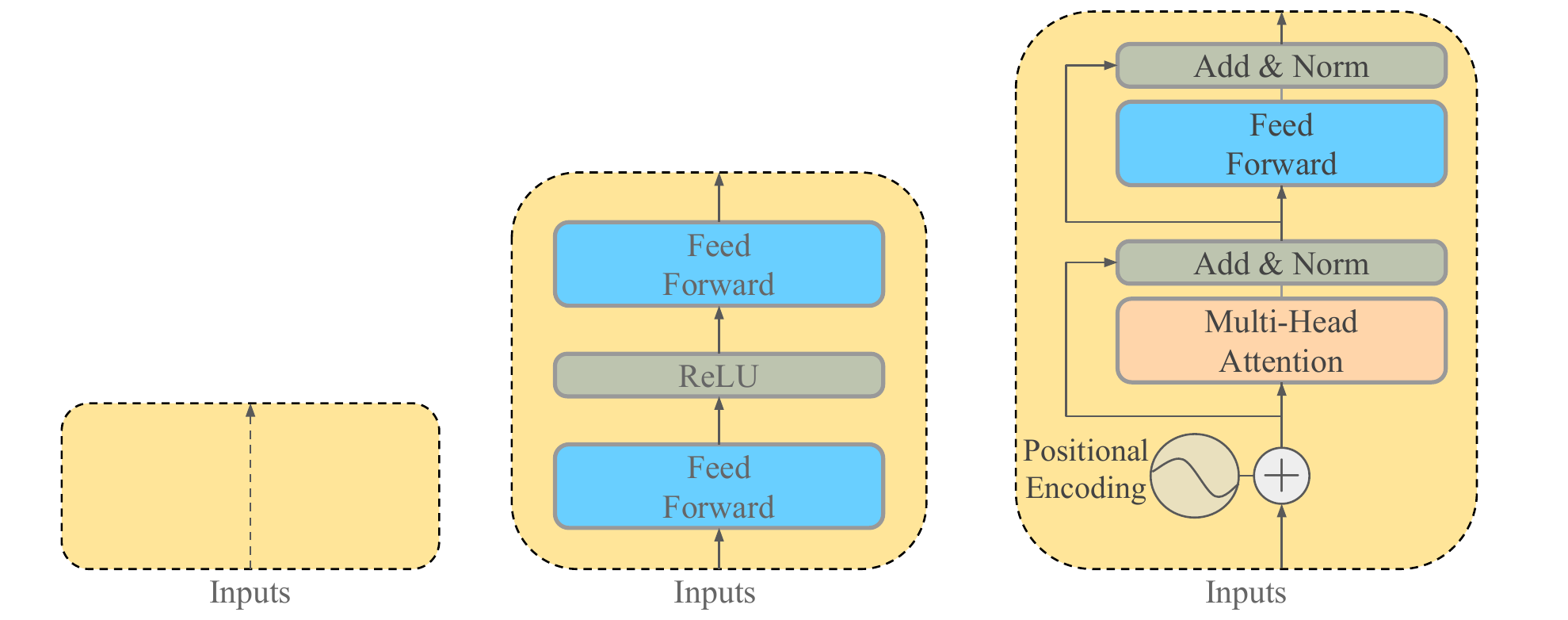}
  \caption{Embedding adapter architectures: Identity (left), \ac{MLP} (middle), and MHA-based (right).}
  \label{fig:adapter}
\end{figure}

We compare two different adapter architectures that are depicted in Figure~\ref{fig:adapter} and contrast them with applying no adapter, which we refer to as the identity function.
The aim is to investigate if one kind of adapter can improve the performance of a Transformer-based \ac{AAC} method.
Moreover, the adapters ensure a match in dimension between the audio embeddings and the internal dimension of the decoder.
It enables us to compare embeddings of different sizes --- from different models --- with a fixed number of decoder parameters.
When not using any adapter, the first decoder layer changes in size depending on the embeddings' dimensionality.

As the first adapter, we employ a two-layer \textbf{\ac{MLP}} with a hidden layer of size 256 and \ac{ReLU} as activation function, as shown in Figure~\ref{fig:adapter}, to compute the adapted representation with weights shared across time.
The second adapter layer is an \textbf{\ac{MHA}} block, followed by a linear layer and a layer normalization process, also known as a Transformer encoder layer~\cite{Vaswani2017AttentionIA}.
This type of network was previously combined with a VGGish embedding model in the context of \ac{AAC}~\cite{koizumi2020audio}.
It complements our decoder in such a way that our model architecture is similar to a full Transformer model for \ac{AAC}~\cite{Koizumi2020TRACKE}.
The \ac{MHA} block employs four attention heads of 128 dimensions.
A linear dimensionality reduction function as described in~\cite{Koizumi2020TRACKE} and a positional encoding precede it.

\subsection{Word embedding models}
\label{ssec:wecomp}
Our first word embedding model is \textbf{word2vec}~\cite{Mikolov2013DistributedRO}, which is based on the skip-gram algorithm.
We use the publicly available model pre-trained on three million words and phrases from Google News.

Our second model is \textbf{GloVe}~\cite{pennington2014glove}, which takes a different approach by learning context information from corpus-level word-word co-occurrence statistics rather than local context windows.
The authors show that GloVe is an improvement over the word2vec algorithm in downstream word analogy and Named Entity Recognition tasks.
We employ the publicly available model trained on the combination of a 2014 Wikipedia copy and the Gigaword 5 corpus~\cite{gigaword5}, which together contain six billion tokens.

The third word embedding model is \textbf{fastText}, which implements several optimizations on top of the word2vec skip-gram algorithm~\cite{mikolov2018advances}.
FastText provides better handling of multi-word phrases, uses a weighted context, and considers subwords (i.e. character n-grams).
We use the publicly available model trained with subword information on the Common Crawl corpus, which contains 600B tokens and is significantly larger than the corpora used for the Glove and word2vec model~\cite{commoncrawl2017}.

We employ \textbf{BERT} as our fourth model, which is a large language model based on the Transformer architecture and can be used as a feature extractor to extract word embeddings%
~\cite{devlin-etal-2019-bert}.
In contrast to models mentioned above, such as word2vec, BERT also takes the context of a token --- the entire sentence --- into account when extracting embeddings, i.e. producing embeddings that are context-sensitive.
We use the BERT\textsubscript{BASE} configuration pre-trained on a Wikipedia copy (2.5B words) and the BookCorpus dataset (800M words)~\cite{zhu2015bookcorpus}.
Different ways to use the layers of BERT as word embeddings have been discussed in the literature, and it is not clear what the best choice is for \ac{AAC}.
We decided to use the penultimate layer as embeddings as this can produce highly contextualized representations that are not too task-specific~\cite{Rogers2021BERTology}.
We extract the word embeddings from an entire caption.
Due to the computational cost of the model, it will not be fine-tuned in our experiments.

Additionally, to explore if pre-trained word embeddings can be helpful, we also adopt randomly initialized word vectors and a \ac{CBOW} word2vec model~\cite{Mikolov2013DistributedRO} trained using the text of the captions in the Clotho training set.
Finally, all word embedding models produce embeddings with $W'=300$ dimensions, except for those from the BERT model with $W'=768$.

\section{Results and Discussion}
\label{sec:results}
In this section, we show and discuss the results of our experiment. 
We organize our discussion around, first, the performance of the different pre-trained audio encoder models.
Second, we discuss the usage of the different audio embedding adapter components. 
Third, we study the performance of different word embedding models and the impact of fine-tuning them.
Finally, we show the potential of computing audio embeddings on overlapped audio frames.

\begin{figure}[t]
  \centering
  \includegraphics[clip,width=\columnwidth]{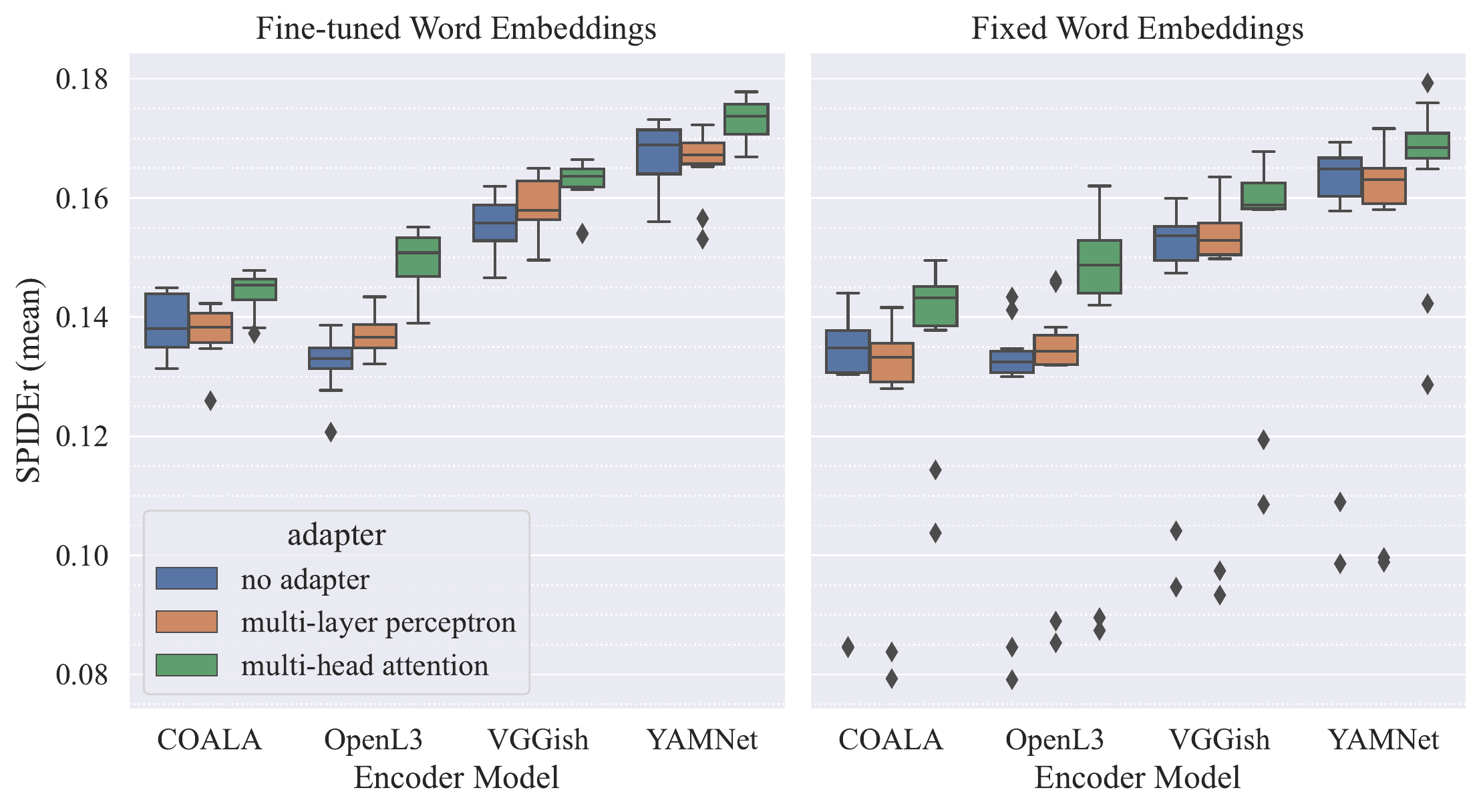}
  \caption{Comparison of SPIDEr score for different encoder models averaged over multiple different experiment settings.}
  \label{fig:sourcemodelcomp}
\end{figure}

Table~\ref{table:bestmodels} lists the optimal settings for each of the audio encoder models that we found, and Figure \ref{fig:sourcemodelcomp} displays box plots of the audio encoder models' performance for each adapter.
The top-performing model in the best overall setting (YAMNet, BERT, \ac{MHA}) achieved a SPIDEr score of 0.1914.
Moreover, we found that YAMNet consistently outperforms the other audio models.
Overall, audio encoder models trained in a supervised classification task using large datasets, YAMNet \& VGGish, are superior to the models that are trained in a self-supervised way or using contrastive learning.
This highlights the potential of using large datasets for pre-training audio encoders in auto-tagging tasks and using them for \ac{AAC}. 

\begin{table}[!t]
\centering
\caption{Top-performing settings for pre-trained encoder models and their SPIDEr score when embeddings are extracted with no or 50\% overlap ($\star$, and $\dag$ respectively).}
\label{table:bestmodels}
\begin{tabular}{lm{1.7cm}lllrr}
& & & \multicolumn{2}{c}{SPIDEr}\\
Encoder & Word\newline embedding        & Adapter   & Mean   & SD\\
\midrule
 COALA\textsuperscript{$\dag$}   & BERT & MHA-based & 0.1495 & 0.0044 \\
 OpenL3\textsuperscript{$\star$} & BERT & MHA-based & 0.1620 & 0.0051 \\
 VGGish\textsuperscript{$\star$} & BERT & MHA-based & 0.1677 & 0.0052 \\
 YAMNet\textsuperscript{$\dag$}  & BERT & MHA-based & 0.1793 & 0.0066 \\
\end{tabular}
\end{table}

Using an \ac{MHA}-based encoder as an adapter on top of the audio embeddings consistently provides the best results (Figure \ref{fig:sourcemodelcomp} and Table \ref{table:bestmodels}), whereas using the \ac{MLP} does not provide any improvement in comparison with no adapter.
Interestingly, the benefit of the \ac{MHA}-based adapter is most prominent for OpenL3, which has been trained in a self-supervised way.
This suggests that the audio embeddings extracted with OpenL3 contain some semantics useful for \ac{AAC} that can be exploited using an adapter such as the \ac{MHA}-based one.
Our results suggest that employing a Transformer-based encoder using positional encoding and \ac{MHA} can process sequences of audio embeddings, leading to better performance in \ac{AAC}, which aligns well with findings from previous works~\cite{Tran2020WaveTransformerAN, Koizumi2020TRACKE}.

\begin{figure}
     \centering
     \includegraphics[clip,width=\columnwidth]{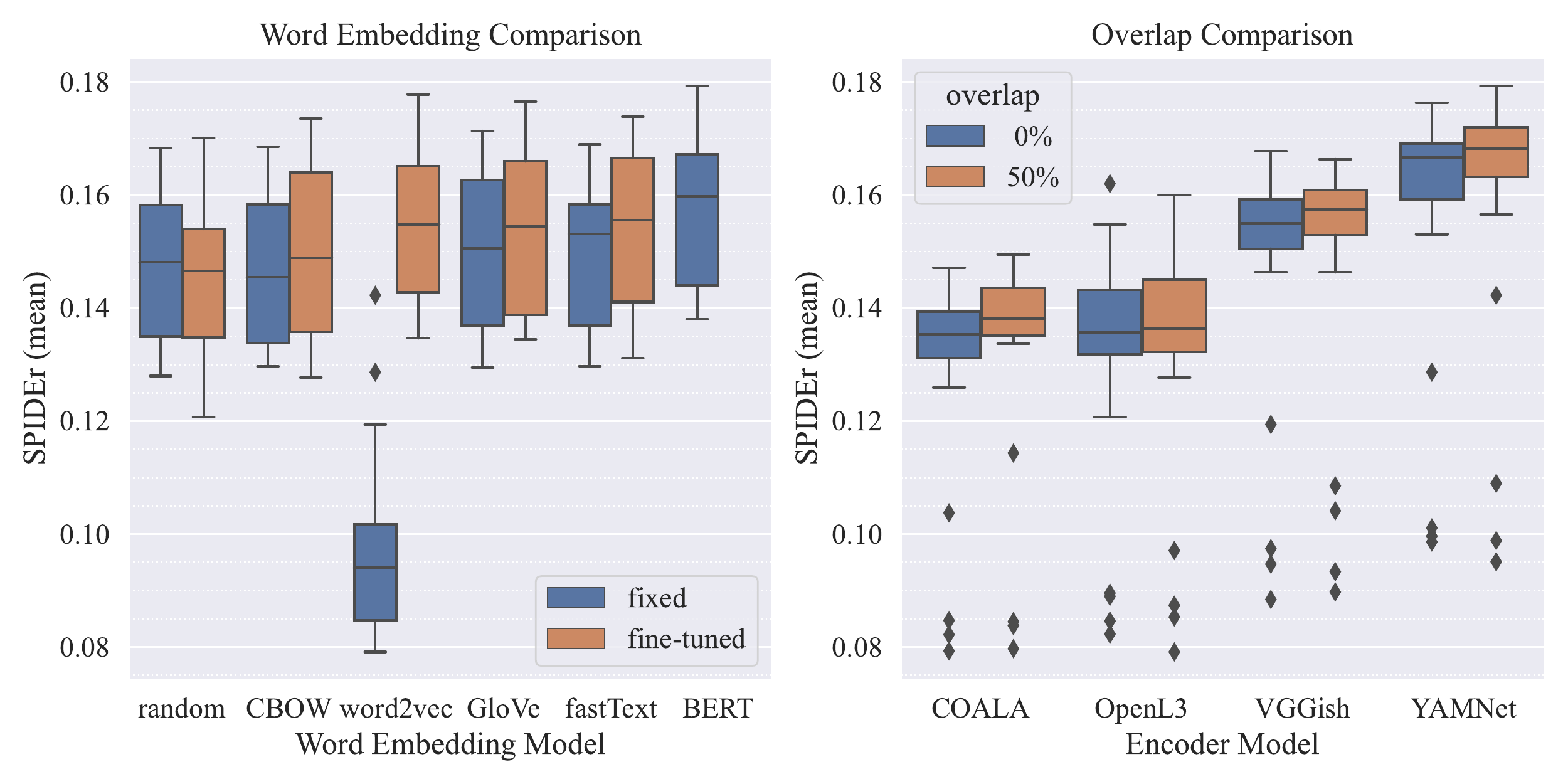}
     \caption{SPIDEr sores for word embedding (left) and audio embedding models (right). Scores are averaged over all combinations is each case.}
     \label{fig:comp}
\end{figure}

The left part of Figure~\ref{fig:comp} reports a performance comparison of the different word embeddings employed in our evaluation.
On average, using the pre-trained BERT model to extract the word embeddings leads to the best performance.
It is worth mentioning that we are using BERT as a fixed external word embedding model instead of using its full capacity, for example, by fine-tuning it for \ac{AAC}.
However, the latter would require much more computational resources (the BERT model has around 110M parameters).

Training word embedding representations from scratch during \ac{AAC} provides already promising results.
By using pre-trained word embeddings, such as word2vec, GloVe, and fastText, we can slightly improve this performance.
Additionally, fine-tuning them can significantly improve their performance (one-sided Wilcoxon signed-rank tests $p<0.001$, for each pre-trained word embedding model).
Interestingly, optimizing the randomly initialized word representations does not improve their performance. 
This highlights the need for pre-trained word representations in the context of our \ac{AAC} task.

The right side of Figure \ref{fig:comp} displays the performance of the audio models with and without overlap when extracting the embeddings.
In particular, we observe that computing the embeddings with 50\% overlap leads to improved performance with two audio encoders.
One-sided Wilcoxon signed-rank tests indicated that this improvement is significant for COALA ($W=84, p<8.08e^{-07}$) and YAMNet ($W=112, p<3.92e^{-06}$).

\acresetall
\section{Conclusion}
\label{sec:conclusions}
In this paper, we conduct a comparative analysis of many off-the-shelf resources from \ac{NLP} and the machine listening field for \ac{AAC}.
The core components of our method are a fixed audio encoder, an audio embedding adapter, and a Transformer-based decoder.
Our results show that YAMNet outclasses the other audio embedding models when used as an encoder.
The performance can be increased for two encoders (COALA \& YAMNet) by computing the embeddings on overlapped frames.
Processing the audio embeddings with a \acl{MHA}-based adapter can increase the performance of our captioning system while using a \acl{MLP} is not different from not using any adapter.
We found that pre-trained word embedding models are a valuable resource for \ac{AAC}, particularly so when fine-tuned during the training.
Using BERT as a fixed embedding extraction model gave the best results.
This result motivates the usage of large pre-trained \ac{NLP} models such as BERT to create better \ac{AAC} methods.

Future work could investigate the impact of the positional encoding in the audio adapter by independently evaluating the \acl{MHA} adapter.
Finally, fine-tuning the audio embedding models has not been studied in this work.
However, it may be an essential technique that can benefit \ac{AAC} approaches, as highlighted by the fact that adding an adaptation model to process the embeddings significantly increases the performance of our system.

\section{ACKNOWLEDGMENT}
\label{sec:ack}
K. Drossos has received funding from the European Union’s Horizon 2020 research and innovation programme under grant agreement No 957337, project MARVEL. 


\fontsize{8}{9}\selectfont
\bibliographystyle{IEEEtran}
\bibliography{references}

\begin{acronym}
  \acro{AAC}{automated audio captioning}
  \acro{ASC}{acoustic scene classification}
  \acro{CBOW}{continuous bag-of-words}
  \acro{CNN}{convolutional neural network}
  \acro{MaL}{machine listening}
  \acro{MHA}{multi-head attention}
  \acro{MLP}{multi-layer perceptron}
  \acro{NLP}{natural language processing}
  \acro{ReLU}{rectified linear unit}
  \acro{RNN}{recurrent neural network}
  \acro{seq2seq}{sequence-to-sequence}
\end{acronym}

\end{document}